\documentclass[runningheads]{llncs}

\usepackage{comment}
\usepackage{subcaption}
\usepackage{wrapfig}
\usepackage{threeparttable}
\usepackage{makecell}
\usepackage{etex}
\usepackage{hyperref}
\usepackage{diagbox}
\usepackage[utf8]{inputenc}
\usepackage[T1]{fontenc}
\usepackage[ngerman,english]{babel}
\usepackage{times}
\usepackage{multirow}
\usepackage[dvipsnames,table,x11names]{xcolor}
\usepackage{booktabs}
\usepackage{paralist}
\usepackage{amssymb, amsfonts}
\usepackage{amsmath}
\usepackage{amsopn}
\usepackage{url}
\usepackage{multirow}
\usepackage{relsize}
\usepackage{xfrac}
\usepackage{pdflscape}
\usepackage{cite}
\usepackage{tabularx}
\usepackage{mdframed}
\usepackage{marginnote}
\usepackage{algorithm}
\usepackage{algorithmicx}
\usepackage[noend]{algpseudocode}
\usepackage{inconsolata}
\usepackage{nicefrac}
\usepackage{xspace}
\usepackage{tabu}
\usepackage{todonotes}
\usepackage{listings}

\hypersetup{
	pdftoolbar=true,        
	pdfmenubar=true,        
	pdffitwindow=false,     
	pdfstartview={FitH},    
	pdftitle={RAG},    
	pdfauthor={anonym},     
	pdfsubject={},   
	pdfcreator={},   
	pdfproducer={}, 
	pdfkeywords={}, 
	pdfnewwindow=true,      
	colorlinks=true,       
	linkcolor=Brown,          
	citecolor=Brown,        
	filecolor=Brown,      
	urlcolor=Brown           
}

\algnewcommand{\LineComment}[1]{\Statex \(\triangleright\)  {\smaller \ttfamily{\textbf{#1}}}}
\algnewcommand{\LineCommentTwo}[2]{\Statex \hspace{#1}\(\triangleright\)  {\smaller \ttfamily{\textbf{#2}}}}

\algrenewcomment[1]{\hfill\(\triangleright\){\smaller \ttfamily{\textbf{#1}}}}

\newcolumntype{R}[1]{>{\raggedleft\let\newline\\\arraybackslash\hspace{0pt}}m{#1}}

\newcommand{\mypar}[1]{\noindent\textbf{#1.}}
\newcommand{\myparit}[1]{\noindent\textit{#1.}}

\begin{document}
\title{Detecting Undesired Process Behavior by Means of Retrieval Augmented Generation}
\titlerunning{Detecting Undesired Behavior with RAG}



%
\author{Michael Grohs\inst{1,2}\orcidID{0000-0003-2658-8992} \and
Adrian Rebmann\inst{2} \orcidID{0000-0001-7009-4637} \and
Jana-Rebecca Rehse\inst{1}\orcidID{0000-0001-5707-6944}}
\authorrunning{M. Grohs et al.}
%
\institute{University of Mannheim, Germany \\
\email{\{michael.grohs,rehse\}@uni-mannheim.de}
  \and
SAP Signavio, Berlin, Germany \\
\email{adrian.rebmann@sap.com}\\
}
\maketitle              
\begin{abstract}
Conformance checking techniques detect undesired process behavior by comparing process executions that are recorded in event logs to desired behavior that is captured in a dedicated process model. 
If such models are not available, conformance checking techniques are not applicable, but organizations might still be interested in detecting undesired behavior in their processes.
To enable this, existing approaches use Large Language Models (LLMs), assuming that they can learn to distinguish desired from undesired behavior through fine-tuning. 
However, fine-tuning is highly resource-intensive and the fine-tuned LLMs often do not generalize well.
To address these limitations, we propose an approach that requires neither a dedicated process model nor resource-intensive fine-tuning to detect undesired process behavior. 
Instead, we use Retrieval Augmented Generation (RAG) to provide an LLM with direct access to a knowledge base that contains both desired and undesired process behavior from other processes, assuming that the LLM can transfer this knowledge to the process at hand. 
Our evaluation shows that our approach outperforms fine-tuned LLMs in detecting undesired behavior, demonstrating that RAG is a viable alternative to resource-intensive fine-tuning, particularly when enriched with relevant context from the event log, such as frequent traces and activities.

\end{abstract}

\section{Introduction}
Conformance checking is a sub-discipline of process mining which compares process executions that are captured as traces recorded in an event log to process behavior captured in a process model \cite{Dunzer_2019SOA_CC}.
As input, conformance checking techniques require an event log and a dedicated process model that explicitly specifies what behavior is desired in the process \cite{carmona2018conformance}.
Based on this, they can identify which traces diverge from the desired behavior and, subsequently, locate the undesired behavior within the trace \cite{carmona2018conformance}.
This undesired behavior has negative impact on any objective of the process, including but not limited to compliance, performance, and outcome. In this paper, we focus on undesired control-flow of processes, i.e., activity executions that are not envisioned.

The applicability of conformance checking hence depends on the availability of high-quality models for the process in question~\cite[p.11]{Mun16fundamentalsCC}. 
However, creating and maintaining such models requires substantial effort and stakeholder involvement \cite{friedrich2011process}, which means that they are often unavailable, incorrect, or out of date \cite{Dunzer_2019SOA_CC}. 
Further, the process at hand might be too complex to be effectively captured in a model~\cite{van2012makes} or the organization might decide not to model its processes to spare itself the efforts \cite{friedrich2011process}. 
In all of these cases, conformance checking is not applicable. Nevertheless, organizations might still want to know whether their processes contain undesired behavior \cite{caspary2023does}.

To alleviate this issue, several approaches have been proposed that detect undesired behavior in event logs without a dedicated process model.
Unsupervised anomaly detection techniques identify statistically infrequent traces, assuming these are undesired \cite{nolle_binet, lahann2022lstm}. However, since low frequency does not always indicate undesirability, some approaches instead leverage the natural language semantics of activities to detect undesired process behavior \cite{van2021natural}.
Recent approaches leverage the natural language understanding capabilities of Large Language Models (LLMs) for this task. The key idea is that fine-tuning an LLM on process behavior enables it to learn how processes function, allowing it to differentiate between desired and undesired behavior \cite{busch2024xsemad, caspary2023does, guan2024dabl}.
However, fine-tuning is highly resource-intensive and the resulting models often do not generalize well, especially to processes that differ considerably from those used for fine-tuning \cite{busch2024xsemad}. 

To overcome these limitations, we propose an approach for detecting undesired process behavior that requires neither a dedicated process model nor resource-intensive fine-tuning. 
Instead, we use Retrieval Augmented Generation (RAG), which builds on the idea that LLMs answer more accurately if they are supplied with explicit task-related knowledge that has not been used in training \cite{lewis2020retrieval}.
Concretely, we provide an LLM with direct access to a knowledge base that captures both desired and undesired behavior of a broad range of processes.
In addition, we provide it with explicit knowledge about the event log at hand in form of frequent traces and activities.
Then, the LLM can use this task-related knowledge to find undesired behavior in previously unseen traces.
In particular, our approach can detect instances of five established patterns of undesired behavior~\cite{hosseinpour2024auditors}: \textit{inserted}, \textit{skipped}, \textit{repeated}, \textit{replaced}, and \textit{swapped} activities (or sequences thereof). 
Thus, our approach can identify more complex instances of undesired behavior compared to existing approaches, which are limited to deviations involving at most two activities \cite{busch2024xsemad, caspary2023does}, even though undesired behavior can span any number of activities.
Our evaluation experiments show that our approach outperforms both general-purpose LLMs and LLMs that have been fine-tuned for detecting undesired process behavior.

The remainder of this paper is structured as follows.
We first discuss related work in \autoref{sec:related}. Then, we present our approach in \autoref{sec:appr} and evaluate it in \autoref{sec:eva}. 
We discuss our findings and conclude in \autoref{sec:conclusion}.

\section{Related Works} 
\label{sec:related}

Our work relates to conformance checking, unsupervised anomaly detection, and semantic anomaly detection. Also, it relates to RAG-usage in other process mining tasks.

\mypar{Conformance Checking} Traditional conformance checking techniques compare event log traces to process models that define desired process behavior \cite{carmona2018conformance}.
They all require a process model as input \cite{Dunzer_2019SOA_CC}. 
State-of-the-art techniques are trace alignments \cite{Dunzer_2019SOA_CC}, which identify inserted and missing activities. 
Based in these alignments, it is also possible to identify high-level patterns of undesired behavior such as repetitions or swaps \cite{garcia_2017_complete,grohs2024beyond}. 
Furthermore, declarative conformance checking techniques check the satisfaction of constraints, identifying violations of temporal patterns like precedence or response \cite{deleoni2012aligning}. Unlike our approach, all corresponding techniques require a process model as input.

\mypar{Unsupervised Anomaly Detection} 
Unsupervised anomaly detection tries to find undesired behavior without relying on a process model, by identifying statistically rare behavior in a given event log. 
Corresponding techniques are, e.g., based on association rule mining \cite{böhmer2020association} or likelihood graphs of traces \cite{böhmer2017multi}.
Deep learning methods like GRU autoencoder neural networks combined with an anomaly threshold \cite{nolle_binet} or LSTM networks \cite{lahann2022lstm} have also been proposed.
All these techniques determine whether a process instance is statistically unlikely. 
However, likeliness does not necessarily correspond to desirability, so infrequency alone does not suffice to detect deviations from desired process behavior. This constitutes a conceptual difference to the goal of our approach.

\mypar{Semantic Anomaly Detection} 
Rather than relying on infrequency, several approaches employ language models, which allows them to consider the natural language semantics of activities to identify undesirable behavior. This has been done on trace-level and activity-level.
On trace-level, semantically anomalous traces can be detected~\cite{rebmann2024evaluating}, but this does not allow for locating undesired activities. 
On activity-level, some approaches identify semantically anomalous ordering relationships between pairs of activities \cite{caspary2023does,rebmann2024evaluating}, but can therefore not detect more complex undesired behavior encompassing more than two activities.
Another approach on activity-level (called \emph{xSemAD}) detects violated declarative constraints by fine-tuning a sequence-to-sequence model~\cite{busch2024xsemad}. 
While constraints explain why certain activity pairs are undesired in traces, they are limited to just two activities. When violations span more than two activities, xSemAD detects multiple individual constraint violations, making it difficult to identify more complex instances of undesired behavior. Furthermore, despite requiring substantial resources for fine-tuning, xSemAD struggles with precision, due to a many false positive violations.
Finally, an activity-level approach, called \emph{DABL}, fine-tunes an LLM to detect insertions, skips, repetitions, and swaps of activities in traces \cite{guan2024dabl}. 
Since the authors have not evaluated their approach regarding its ability to detect which activities deviate, the efficacy and efficiency of such a fine-tuning strategy remains uncertain. 
In contrast to xSemAD and DABL, our approach does not require any fine-tuning.

\mypar{Retrieval Augmented Generation} 
The idea behind RAG is to provide an LLM with relevant information that it has not seen during training. This is particularly helpful for knowledge-intensive tasks \cite{lewis2020retrieval}. 
The relevant information is \textbf{R}etrieved from a knowledge base that contains documents, textual reports, or other related data objects, then \textbf{A}ugments an LLM prompt, which eventually \textbf{G}enerates the desired output \cite{lewis2020retrieval}. 
Compared to fine-tuning the LLM with the relevant information directly, this is less computationally expensive and can lead to less overfitting on the training data \cite{lewis2020retrieval}.
In process mining, RAG has been rarely used so far. 
One application uses it in a conversational agent based on process model retrieval and fine-tuning to support human decisions \cite{bernardi2024conversing}. 
Another uses it to support the deployment of processes and their operations \cite{monti2024nl2processops}. 
Our approach is the first to use RAG to detect deviations from desired behavior.

\section{Approach} 
\label{sec:appr}

This section presents our approach for detecting undesired behavior in an event log by means of RAG. 
\autoref{sec:overall} gives an overview of the approach, before \autoref{sec:offline} and \autoref{sec:online} detail its components and phases.

\subsection{Overview}
\label{sec:overall}
In this section, we give an overview of our approach's input, design, and output. 

\mypar{Input} As input, our approach takes a process model collection $\mathcal{M}$ and an event log $L$. 
To define them, let $\mathcal{A}$ be the universe of possible activities that can be performed organizational processes. 
A process model $M$ is a finite set of desired execution sequences  over a set of activities $A \subseteq \mathcal{A}$, where each sequence $s \in M$, with $s=\langle a_1,\dots,a_n\rangle$ and $a_i \in \mathcal{A}$, should lead the process from an initial to a final state. A process model collection $\mathcal{M}$ is a finite set of process models. 
Further, focusing on the control-flow of a process, we define a trace $t=\langle a_1,\dots,a_n\rangle$, with $a_i \in \mathcal{A}$, as a sequence of activities that have been executed during the execution of a single instance of an organizational process.  
An event log $L$ is a finite multi-set of such traces.

\begin{figure}[t]
     \centering
         \includegraphics[width=.9\textwidth]{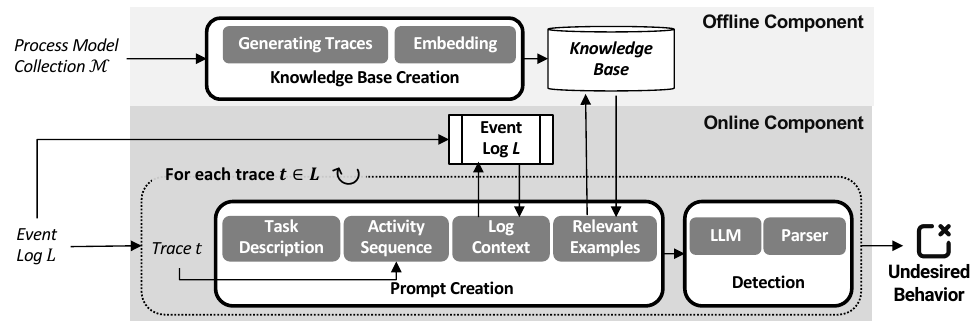}
        \caption{Overview of our approach for detecting undesired behavior}
        \label{fig:approach}
        \vspace{-.5em}
\end{figure}

\mypar{Design} The general idea of our approach is to enable an LLM to detect undesired behavior in an event log by providing it with desired and undesired behavior from similar processes as well as explicit information about that log. To achieve that, our approach consists of an offline and an online component (cf. \autoref{fig:approach}). 
\begin{compactenum}[(1)]
\item \emph{Offline Component} (\autoref{sec:offline}): Given a process model collection $\mathcal{M}$, the offline component populates a knowledge base of desired and undesired process behavior. This behavior can be accessed by an LLM by means of RAG and used to detect undesired behavior in a process that is not contained in the knowledge base.
\item \emph{Online Component} (\autoref{sec:online}): The online component detects undesired behavior in event log $L$, analyzing each trace $t \in L$ individually. 
For that, it retrieves similar behavior from the knowledge base and extracts relevant context from $L$ in form of frequent behavior. Then, it combines retrieved behavior, extracted log context, and a task description to prompt an LLM which identifies any undesired parts of $t$.
\end{compactenum}

\mypar{Output} Our approach outputs undesired behavior per trace $t$ from event log $L$. To characterize this undesired behavior, we employ a set of five established patterns, so-called deviation patterns~\cite{hosseinpour2024auditors}: \textit{inserted}, \textit{skipped}, \textit{repeated}, \textit{replaced}, and \textit{swapped} activities.\footnote{Note that the original source~\cite{hosseinpour2024auditors} also proposes a \textit{loop} pattern, but in the context of undesired behavior, this is simply a specific version of the \textit{repeated} pattern.} 
These patterns account for undesired behavior that spans any number of activities, thus providing a more comprehensive view of deviations.
In particular, each pattern relates to a specific type of undesired fragment, i.e., one or multiple consecutive activities:
\begin{compactitem}[$\bullet$]
\item The \textit{inserted} pattern indicates that a trace contains a fragment which was not supposed to occur at that point in the trace.

\item The \textit{skipped} pattern indicates that a trace misses a fragment which was supposed to occur at some point in the trace.

\item The \textit{repeated} pattern indicates that a fragment is re-executed in a trace although only the first execution was desired.

\item The \textit{replaced} pattern indicates that one fragment in a trace was executed although a different fragment was supposed to occur.

\item The \textit{swapped} pattern indicates that two fragments of a trace were performed in the wrong order.
\end{compactitem}
If a trace $t$ is found to be undesired, our approach outputs a set of instantiations of these deviation patterns.
If $t$ is found to be desired, the output contains no behavior for $t$.

\subsection{Offline Component}\label{sec:offline}
The offline component populates a knowledge base of desired and undesired behavior, which can be accessed by the online component, allowing an LLM to use this knowledge to detect undesired behavior in previously unseen processes. 
The idea is that the LLM can better detect the five types of deviation patterns when having access to traces from similar processes with known patterns of these types. Complementarily, the LLM might better identify desired behavior when exposed to traces from similar processes that are known to be desired. 
To populate the knowledge base, 
our approach performs two steps. First, it generates traces for each model  $M \in \mathcal{M}$, some of which conform to the model (desired behavior) whereas others are enriched with known deviations from the model (undesired behavior). Second, it transforms the traces into vector representations to allow the online component to retrieve them. We next elaborate on these steps.

\mypar{Generating Traces}
To generate the traces, we use the desired executions defined in $\mathcal{M}$ and additionally create undesired executions by adding noise to a share of them. First, we obtain the desired behavior by creating traces that correspond to the activity sequences captured in all $M \in \mathcal{M}$, resulting in one event log $L^M$ per model.\footnote{Note that we limit the number of loop executions to two.}
Then, we add instantiations of the five deviation pattern types, referred to as \emph{deviations}, to a subset of the traces in $L^M$, aiming for a balance between conforming and deviating traces. These deviations correspond to known undesired behavior.  
The result is a collection of conforming and deviating traces from all $M \in \mathcal{M}$, including information about which (if any) deviations occur where in these traces.

\mypar{Embedding}
The next step in the offline component involves creating numerical representations (\emph{embeddings}) of the generated traces.
These embeddings are essential for quantifying the similarity between individual traces during retrieval from the knowledge base.
To ensure that the similarities factor in both the natural language semantics of the individual activities as well as their ordering relations in the trace, we represent each trace as a sentence (a single string that concatenates the activities). We then employ a pre-trained language model that was specifically designed to capture the meaning of entire sentences to obtain one embedding per trace \cite{nussbaum2024nomic}.
We then store these embeddings, along with any known deviations in the respective trace, in the knowledge base. 
This knowledge base is then used by the online component to search for the most similar embeddings and retrieve them along with its known deviations, as we explain next.

\subsection{Online Component}\label{sec:online}
The online component takes an event log $L$ as input and then performs two steps for each trace $t \in L$. In the \emph{prompt creation} step, it populates a prompt template with process knowledge. The resulting prompt is then used in the \emph{detection} step so that a general-purpose LLM can use it to detect undesired behavior in the trace.

\begin{wrapfigure}{R}{5cm}
     \centering
         \includegraphics[width=.41\textwidth]{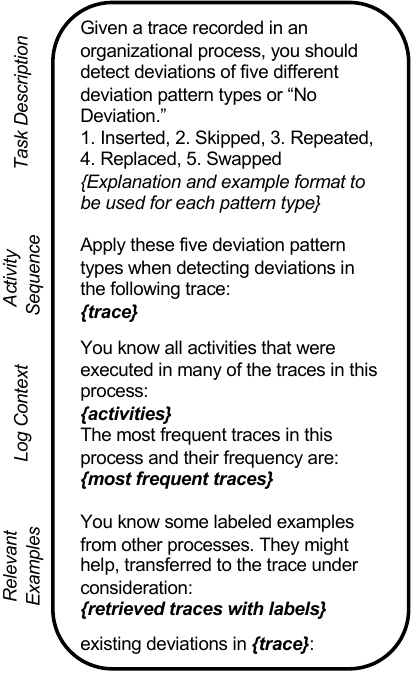}
        \caption{Excerpt from prompt template. Boldness indicates dynamic parts that are adjusted per trace.}
        \label{fig:prompt}
        \vspace{-2.5em}
\end{wrapfigure}

\mypar{Prompt Creation}
The first step of the online component populates a prompt template for the task to detect undesired behavior in a trace $t \in L$. In addition to a textual task description, the template includes the activity sequence of $t$, explicit log context from $L$, and relevant example traces retrieved from the knowledge base. Except for the task description, all other parts are added to the template dynamically for each trace.
In the following, we elaborate on the parts of the prompt template, illustrated in \autoref{fig:prompt}.\footnote{The complete prompt template can be found in our repository: \url{https://doi.org/10.6084/m9.figshare.29125898}}

\myparit{Task Description} 
This first part introduces and explains the five deviation pattern types to be detected. For example, it states that the \textit{inserted} pattern refers to a fragment which was not supposed to occur at a certain point in the trace. We specify that any detected undesired behavior in $t$ must be an instance of one of these pattern types, affecting specific activity fragments---either a single activity or multiple consecutive ones. 
Finally, it asks  to output ``No Deviation.'' if no undesired behavior is found.
In the following, we describe the dynamic parts of the prompt in detail.

\myparit{Activity Sequence} This part is dynamic and corresponds to the sequence of activities in $t$, separated by commas. The prompt states that the previously defined patterns should be applied to this particular activity sequence. 

\myparit{Log Context} This part contains information from $L$ that can be used to reason about undesired behavior in $t$. Concretely, we include activities that were executed in many traces in $L$, giving the LLM an idea of what other important activities could have happened in $t$. Further, we include the most frequent traces from $L$, giving it an idea of what order of activities was often executed and indicating that these traces can be considered a reference for desired behavior. 

\myparit{Relevant Examples} This final part contains traces from the knowledge base that are similar to $t$ plus known deviations in them. To retrieve these traces, we create an embedding of $t$ with the same procedure as in the offline component. Then, we calculate the cosine similarity of all traces in the knowledge base and retrieve a configurable number of traces with the highest similarity to $t$ as well as all known deviations (if any) in them. 
These can be used to generalize and transfer (un)desired behavior to $t$. 


\mypar{Detection}\label{sec:llm}
The second step of the online component detects undesired behavior in $t$ by feeding the prompt into an LLM and parsing the output into a structured format.

\myparit{LLM} We use a general-purpose LLM, designed to generate accurate responses to a wide range of questions. This LLM is not fine-tuned for the specific task at hand, but it has the potential to transfer its general natural language understanding capabilities for this purpose. 
We want to enable this transfer by providing the LLM with the prompt from the previous step, which contains explicit information from the log as well as examples from other processes that illustrate desired and undesired behavior (retrieved from our populated knowledge base). 
The output of the LLM is parsed in the next step.

\begin{figure}[b]
     \centering
         \includegraphics[width=\textwidth]{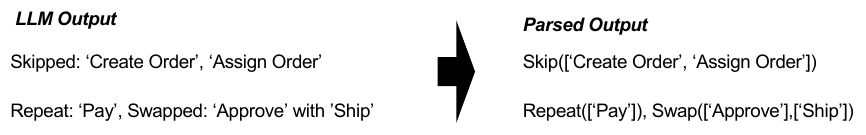}
         \vspace{-1em}
        \caption{Exemplary parsed output}
        \label{fig:parser}
\end{figure}

\myparit{Parser} Finally, we parse the output string that the LLM generated into in a standardized output format. This is necessary because the output of the LLM is not guaranteed to be structured, leading to problems when the output should be used in downstream analyses. 
An example of this standardized format is shown in \autoref{fig:parser}.
To this end, we define a data model for the output of our approach in which we represent each deviation pattern type as a separate entity type. The entity types can be instantiated with trace fragments that are affected by a respective deviation.\footnote{The complete data model can be found in our repository.}
For example, skipped activities should be parsed into the \texttt{Skip()} type that contains a list of the activities in the affected fragment. Swapped and replaced patterns affect two fragments, so they contain two distinct lists of activities (e.g., \texttt{Swap(['Approve'], ['Ship'])}).
Data model and the LLM output string from the previous step are handed back to LLM, which is tasked to parse the string into an instantiation of the data model, enforcing that the defined format is returned. 

Our approach outputs one instantiation of the data model per trace $t \in L$. In case no deviations were found in $t$, the output does not contain any instantiated deviation.

\section{Evaluation}\label{sec:eva}

In this section, we evaluate the capability of our approach to detect undesired behavior, comparing it to four baseline approaches. 
In the following, we first describe the data collection (\autoref{sec:data}) and the experimental setup (\autoref{sec:setup}). Then, we present the results of our experiments in terms of accuracy, robustness, and efficiency, demonstrating that our approach outperforms four baseline approaches in these aspects (\autoref{sec:results}). 
Finally, we showcase the practical applicability of our approach, illustrating that it can identify undesired behavior in a real-life event log (\autoref{sec:eva_real}).
Following the guidelines for LLM-based process analysis from~\cite{estrada2024mapping}, our Python implementation, sources, model configurations, process stages, and outcomes are available in our repository\footnote{\url{https://doi.org/10.6084/m9.figshare.29125898}}. 


\subsection{Data Collection}
\label{sec:data}
To evaluate our approach, we require (i) a process model collection $\mathcal{M}$ to populate the knowledge base in the offline component and (ii) event logs with known deviations of the five deviation pattern types to assess performance in a controlled setting.

\mypar{Process Model Collection}
As process model collection, we use part of the SAP Signavio Academic Models (SAP-SAM)~\cite{sola2022sap} dataset which is the largest publicly available collection to date. This model collection includes very diverse processes, ranging from business processes to education processes. From SAP-SAM, we select English models that meet specific requirements to ensure that the corpus includes a wide variety of valid process behavior. In particular, we require that a model can be transformed into a sound workflow net to allow for a playout of the model and that no two models have the same activity set to ensure the exclusion of duplicates. Last, we require a model to contain at least five different activities to ensure a minimum level of process complexity. The result is a collection of 7,791 process models.

We split these models into two sets, \textit{SAP-SAM-train}, encompassing 80\% (6,232) of the models, and \textit{SAP-SAM-test}, including the remaining 20\% (1,559). 
The former becomes the process model collection $\mathcal{M}$ that we use to populate the knowledge base, whereas the latter is used to generate event logs for testing, as explained below.

\mypar{Event Logs for Test Data} 
We use three sources of test data: (1) the process models in SAP-SAM-test, (2) a smaller, proprietary collection of 396 process models (PROP), and (3) a manually created event log with known deviations \cite{hosseinpour2024auditors}. (2) contains diverse process models from an internal collection, which therefore passed more thorough quality checks than those in SAP-SAM. (3) is a concrete log which contains manifold deviation patterns. By definition, this log cannot be found in the SAP-SAM data, which only consists of process models.

For (1) and (2), we synthetically create event logs with known deviations as ground truth. 
Concretely, we obtain an event log $L^M$ per model $M$ by creating one trace in the log for each sequence $s \in M$. 
If $M$ includes less than 100 sequences, we ensure a minimum number of desired traces per log by randomly duplicating traces in $L^M$ until a size of 100 is reached.\footnote{We note that most models in (1) and all models in  (2) do not contain more than 100 sequences.}
Then, we add undesired behavior to 55\% of the traces to create balanced data. 
For each deviating trace, we add up to three deviations of a randomly chosen pattern type and affected fragments of one to three random activities. 
Then, we verify that the deviation with all its activities is an actual deviation as defined by $M$, and otherwise disregard the deviation and select another until we find a valid one.\footnote{We limit the number of retries to 10 per trace to guarantee termination and randomly select the deviation type again for each retry.} 
Through that, we obtain an event log with known deviations per process model that serves as ground truth.

Last, we also use an event log (3) from \cite{hosseinpour2024auditors} as test data in addition to the synthetically generated event logs. This log contains manually crafted and hence more realistic deviations.
Concretely, we draw on a purchase-to-pay process (P2P) created for the (manual) identification of deviation patterns in traces~\cite{hosseinpour2024auditors}. For this process, the authors created a set of 18 deviating traces and a corresponding ground truth that specifies which deviation patterns of the five types exist in the traces. In addition, the authors defined which traces are desired according to a process model. Combining both desired and deviating traces, we obtain the event log for P2P and use this log as test data.

\autoref{tab:descriptive} shows descriptive statistics of the generated event logs for SAP-SAM-test and PROP as well as statistics for P2P. Note that the deviation patterns are slightly imbalanced, since randomly generated replacements and swaps may not constitute undesired behavior according to the process model, meaning that they are disregarded and another deviation type is chosen. 

\begin{table}[t]
\caption{Descriptive statistics for used event logs.} \label{tab:descriptive}
\resizebox{\textwidth}{!}{
\begin{tabular}{l>{\centering\arraybackslash}m{1.0cm}>{\centering\arraybackslash}m{1.1cm}>{\centering\arraybackslash}m{1.5cm}>{\centering\arraybackslash}m{1.1cm}>{\centering\arraybackslash}m{1.5cm}>{\centering\arraybackslash}m{1.0cm}>{\centering\arraybackslash}m{1.0cm}>{\centering\arraybackslash}m{1.0cm}>{\centering\arraybackslash}m{1.0cm}>{\centering\arraybackslash}m{1.0cm}}
\toprule
        & \multirow{2}{*}{\makecell{Event \\ Logs}} & \multicolumn{3}{c}{Number of Traces}  & \multirow{2}{*}{\makecell{Avg. Devs. \\ per Trace}} & \multicolumn{5}{c}{Number of Deviation Patterns}  \\ \cline{3-5} \cline{7-11}
        && total & conforming& deviating && insert & skip & repeat & replace & swap \\ \midrule
SAP-SAM-test & 1,559 & 180,077 & 82,990 & 97,087 & 0.76 & 34,201 & 28965 & 31,536 & 21,233 & 21,534   \\
PROP    & 396 & 39,600 & 20,402 & 19,198 & 0.66 & 6,702  & 4,073  & 5,837  & 4,749   & 4,666  \\
P2P \cite{hosseinpour2024auditors}    & 1 & 58 & 40 & 18 & 0.51 & 6 & 5 & 12 & 3 & 4  \\ 
\bottomrule
\end{tabular}}
\end{table}

\subsection{Experimental Setup}\label{sec:setup}
In this section, we describe details on the implementation of our approach, the optimization of hyperparameters, evaluation metrics, and used baselines. 

\mypar{Implementation} 
We describe the configuration of the knowledge base population, the selection of language models and output parser, and the execution environment.

\myparit{Knowledge Base Population} To populate the knowledge base in the offline component, we use the process model collection SAP-SAM-train. 
For the generation of traces, we follow the same procedure as for the generation of test data from process models, described in the previous section. 
The resulting traces are embedded and stored in the knowledge base along with their known deviations, as described in \autoref{sec:offline}.
As a result, we obtain a knowledge base with embeddings of 354,127 desired traces and 349,487 undesired traces, which contain 120,827 insertions, 99,133 skips, 117,627 repetitions, 71,738 replacements, and 77,508 swaps.

\myparit{Language Models and Output Parser} We select a pre-trained embedding LLM in the offline component, as well as a general-purpose LLM and a parser in the online component of our approach, configuring seeds for reproducibility. 
As pre-trained embedding LLM, our approach uses Nomic Embed \cite{nussbaum2024nomic}. This model has been shown to capture long context dependencies which can be of high relevance for processes with long term dependencies and also promises computational efficiency \cite{nussbaum2024nomic}. 
As general-purpose LLM, we select Gemma 2 in its 9B version \cite{team2024gemma}. This specific LLM is chosen due to its practical size in combination with an open-sourced API and promising capabilities as shown in multiple benchmarks \cite{team2024gemma}.
For facilitating the parsing of the LLM's output, we employ the BAML API,\footnote{\url{https://docs.boundaryml.com/ref/overview}} an open-source library that allows to define an output schema for the LLM to adhere to. This schema is provided as described in \autoref{sec:online}.  

\myparit{Environment} We conducted experiments on a machine with 256GB of RAM and an NVIDIA A10G Tensor Core GPU with 24 GB of GPU memory.

\mypar{Hyperparameter Optimization} We conducted a grid search to identify the most promising hyperparameters for our approach. To this end, we used 100 process models that are not part of the test set.
In particular, we optimized the following three hyperparameters, with the best performing one highlighted in bold:
\begin{compactenum}[(1)]
\item Number of most frequent traces to add to prompt: [0,1,2,\textbf{3},4,5]
\item Number of retrieved relevant trace examples to add to prompt: [0,3,\textbf{5},8,10,15]
\item Relative frequency of activities (in traces) included in the prompt: [0\%, \textbf{10\%}, 20\%]
\end{compactenum}

\mypar{Metrics}
We measure the capability to detect undesired behavior using precision, recall, and F1-score (\autoref{eq:prec_rec} \hyperref[eq:prec_rec]{a.-c.}). 
If a trace contains a certain deviation, it is classified as a true positive (TP) if the complete fragment of the deviation is identified correctly and as a false negative (FN) otherwise. If a deviation is detected but does not exist in the ground truth, it is a false positive (FP).
In addition, detected deviations might be only partially correct when only parts of the affected fragment are identified. We account for that by assigning partial TPs, FPs, and FNs on activity level.
For example, if one deviation consists of three skipped activities but only two of them are detected, TP is increased by $\frac{2}{3}$ and FN is increased by $\frac{1}{3}$.
Further, pattern types can be interpreted differently since all five used types can be described as (a combination of) insertions and skips. In particular, repetitions are a special case of insertions, replacements consist of a skip (the replaced activities) and an insertion (the replacing activities), and swaps indicate that activities are skipped at one point of the trace and inserted at another point. We account for this by also granting (partial) TPs when an alternative but still (partially) correct interpretation of the deviation patterns is detected. For example, if the approach returns a skip of the replaced activities along with an insertion of the replacing activities instead of a direct replacement, we still consider it correct.

In addition to the capability to detect the five pattern types, we also measure the ability to detect \textit{conform} traces. For that, a trace is a TP if it does not contain any deviations in the ground truth and no deviations are detected. It is a FN if there are deviations in the ground truth but none are detected and it is a FP in the opposite case.
\begin{equation}
\label{eq:prec_rec}
    a. \ \mathit{Prec.} = \frac{TP}{TP+FP} \hspace{1em} b. \ \mathit{Rec.}  = \frac{TP}{TP+FN} \hspace{1em}
    c. \ \mathit{F1} = \frac{2 \times \mathit{Prec.} \times \mathit{Rec.}}{\mathit{Prec.} + \mathit{Rec.}}
\end{equation}

\mypar{Baselines}
To put the performance of our approach into context, we compare it to approaches that employ expensive task-specific fine-tuning instead of RAG as well as to approaches that prompt general purpose (\emph{vanilla}) LLMs without RAG.

\myparit{Fine-Tuned Approaches} We compare against two approaches that use fine-tuned LLMs.
\begin{compactenum}[(a)]
    \item \textit{xSemAD \cite{busch2024xsemad}}: Our first baseline is the state-of-the-art approach for semantic anomaly detection. xSemAD detects undesired behavior in event logs by learning constraints from a process model collection and checking whether these constraints hold in the logs. This learning involves fine-tuning of a sequence-to-sequence model. In contrast to our work, xSemAD does not detect deviation patterns but constraint violations. Although this is a different representation, we can compare it to our approach by interpreting both patterns and constraint violations on activity level. All constraint violations indicate either a skipped or inserted activity. Similarly, all deviation patterns can be seen as inserted and skipped activities, as outlined above. Thus, we transform all deviations in the test data to (combinations of) insertions and skips and only evaluate those. We report the performance of xSemAD with the required threshold parameter $\theta$ that produced the best results for the set of process models we used for hyperparameter optimization.
    \item \textit{DABL \cite{guan2024dabl}}: To relate our approach to a fine-tuning strategy that detects the same deviation patterns as our approach, we compare it to DABL~\cite{guan2024dabl}. The idea behind DABL is to fine-tune Llama 2 (13B version) on the same task as our approach with the difference that replacements are not included. Thus, we treat them as a combination of insertions and skips, as done for xSemAD.
\end{compactenum}

\noindent
Both xSemAD and DABL require fine-tuning of an LLM.
We use the fine-tuned LLMs provided by the authors of both approaches since they are publicly available. This ensures no diversions from the intended approaches. Both relied on the SAP-SAM data set for fine-tuning, using a significantly higher number of process models than is contained in SAP-SAM-train. This minimizes the chance that our approach has access to information which is unknown to xSemAD or DABL. Rather, it is likely that both fine-tuned LLMs have seen parts of our test set SAP-SAM-test during training, which is why the performance of xSemAD and DABL for SAP-SAM can be seen as an upper bound.

\myparit{Vanilla LLM} We compare against two approaches solely based on Vanilla LLMs.
\begin{compactenum}[(c)]
    \item \textit{No Log Context}: We compare our approach to an LLM without any access to explicit knowledge of the event log at hand. This baseline follows a so-called zero-shot strategy, which means that it only provides the task description to the LLM, without log context and retrieved examples in the prompt.
\end{compactenum}
\begin{compactenum}[(d)]
    \item \textit{Log Context}: To show the value of RAG for our approach, we compare it to a Vanilla LLM with only access to the log context contained in our prompt. Thus, the LLM does not receive examples retrieved from the knowledge base. This baseline received the same hyperparameter optimization as our approach.
\end{compactenum}

\subsection{Experimental Results}\label{sec:results}
This section presents the results of our evaluation experiments. We report on the accuracy, robustness, and computational efficiency in detecting undesired behavior. 

\mypar{Accuracy}
\autoref{tab:eva_results} shows precision, recall, and F1 scores for the different test data sets and compares the performance of our approach to the four baselines. The numbers for SAP-SAM-test and PROP are averages across all logs. The results for each individual log can be found in our repository. We first discuss overall results before shifting the focus to the comparison against the individual baselines.

\begin{table}[!b]
\centering
\caption{Precision, recall, and F1 for used datasets.  Boldness indicates best score.}
\label{tab:eva_results}
\resizebox{\textwidth}{!}{
\begin{tabular}{>{\centering\arraybackslash}m{.85cm}l>{\centering\arraybackslash}m{.65cm}>{\centering\arraybackslash}m{.65cm}>{\centering\arraybackslash}m{.65cm}m{0.05cm}>{\centering\arraybackslash}m{.65cm}>{\centering\arraybackslash}m{.65cm}>{\centering\arraybackslash}m{.65cm}m{0.05cm}>{\centering\arraybackslash}m{.65cm}>{\centering\arraybackslash}m{.65cm}>{\centering\arraybackslash}m{.65cm}m{0.05cm}>{\centering\arraybackslash}m{.65cm}>{\centering\arraybackslash}m{.65cm}>{\centering\arraybackslash}m{.65cm}m{0.05cm}>{\centering\arraybackslash}m{.65cm}>{\centering\arraybackslash}m{.65cm}>{\centering\arraybackslash}m{.65cm}}
\toprule
 \multirow{3}{*}{\textbf{Data}} & \multirow{3}{*}{\textbf{Pattern}} & \multicolumn{7}{c}{\textbf{Fine-Tuned Approaches}} & &
 \multicolumn{7}{c}{\textbf{Vanilla LLM}} & &
 \multicolumn{3}{c}{\multirow{2}{*}{\textbf{Our Approach}}} \\ 
 & & \multicolumn{3}{c}{\textbf{xSemAD}\cite{busch2024xsemad}} & & \multicolumn{3}{c}{\textbf{DABL}\cite{guan2024dabl}} &  & \multicolumn{3}{c}{\textbf{No Log Context}}&  & \multicolumn{3}{c}{\textbf{Log Context}} &  & \multicolumn{3}{c}{}  \\
 &  & \textbf{Prec.}& \textbf{Rec.} & \textbf{F1} & & \textbf{Prec.} & \textbf{Rec.} & \textbf{F1} & & \textbf{Prec.} & \textbf{Rec.} & \textbf{F1} & & \textbf{Prec.} & \textbf{Rec.} & \textbf{F1} & & \textbf{Prec.} & \textbf{Rec.} & \textbf{F1} \\ \midrule

& inserted &  0.22 & 0.44 & 0.23 & & \textbf{0.78} & 0.45 & 0.57 & & 0.69 & 0.16 & 0.24 & & \textbf{0.78} & 0.81 & \textbf{0.79} & & 0.77 & \textbf{0.82} & \textbf{0.79} \\
& skipped   & 0.14 & 0.69 & 0.20 & &  0.15 & 0.02 & 0.03 & & 0.02 &  0.00 & 0.00 & & \textbf{0.68} & 0.30 & \textbf{0.39} & & 0.58 & \textbf{0.32} & 0.38  \\
& repeated & - & - & - & & \textbf{0.92} & 0.48 & 0.61 & & 0.83 & 0.78 & 0.78 & & 0.84 & 0.78 & 0.80 & & 0.88 & \textbf{0.81} & \textbf{0.83} \\
& replaced & - & - & - &&  - & - & - & & 0.28 & 0.06 & 0.09 & & \textbf{0.68} & 0.42 & 0.51 &  & 0.64 & \textbf{0.44}& \textbf{0.52}  \\
& swapped   & -  & -  & -&  & 0.28  & \textbf{0.33} & \textbf{0.28} & & 0.48  & 0.06  & 0.09 & & 0.58 & 0.13 & 0.19  & & \textbf{0.60} & 0.16 & 0.23  \\ \cline{2-21}
 \parbox[t]{4mm}{\multirow{-6}{*}{\rotatebox[origin=c]{90}{\textbf{SAP-SAM-test}}}} & conforming & 0.11  & 0.11 & 0.11 &  & 0.63 & 0.72 & 0.66 & & 0.54 & 0.82 & 0.64 & & \textbf{0.89} & 0.94 & 0.90 & & 0.88 & \textbf{0.95} & \textbf{0.91} \\\midrule

& inserted & 0.18  & 0.41 & 0.17 & & \textbf{0.85}& 0.35 & 0.49 & & 0.70 & 0.06 & 0.11 & & 0.74 & 0.61 & 0.64 & &0.80 & \textbf{0.79} & \textbf{0.79} \\
& skipped   & 0.12  & \textbf{0.66} & 0.17 & & 0.11  & 0.01 & 0.02 & & 0.00 & 0.00 & 0.00 & & \textbf{0.60} & 0.33 & \textbf{0.39} & & 0.55 & 0.31 & 0.36 \\
 & repeated & -  & - & - & & \textbf{0.95}  & 0.48 & 0.61 & & 0.82 & 0.76 & 0.76 & & 0.84 & 0.77 & 0.78 & & 0.87 & \textbf{0.79} & \textbf{0.80} \\
 & replaced   & -  & - & - & & -  & - & - & & 0.32 & 0.02 & 0.04 & & 0.67 & 0.37 & 0.47 & & \textbf{0.70} & \textbf{0.46} & \textbf{0.55} \\
 & swapped   & -  & - & - & & 0.20  & \textbf{0.26} & \textbf{0.19} & & 0.41 & 0.04 & 0.07  & & 0.61 & 0.11 & 0.17 & & \textbf{0.63} & 0.13 & \textbf{0.19} \\ \cline{2-21}
 \parbox[c]{4mm}{\multirow{-6}{*}{\rotatebox[origin=c]{90}{\textbf{PROP}}}}   & conforming & 0.05  & 0.07 & 0.06 & & 0.50  & 0.67 & 0.57&  & 0.56 & 0.86 & 0.66 & & \textbf{0.81} & 0.91 & 0.85 &  & \textbf{0.81} & \textbf{0.92} & \textbf{0.86} \\ \midrule

 & inserted & 0.14  & \textbf{0.69} & 0.24 & & 0.83  & 0.35 & 0.50 & & \textbf{1.00} & 0.30 & 0.46 & & \textbf{1.00} & 0.30 & 0.46 & & \textbf{1.00} & 0.60 & \textbf{0.75} \\
 & skipped   & 0.05  & \textbf{0.70} & 0.10 & & \textbf{1.00}  & 0.10 & 0.18 & & 0.00 & 0.00 & 0.00 & & 0.40 & 0.43 & 0.41 & & \textbf{1.00} & 0.57 & \textbf{0.73} \\
 & repeated & -  & - & - & & 0.92  & 0.50 & 0.65 & & 0.95 & 0.54 & 0.69 & & \textbf{1.00} & 0.79 & 0.88 & & 0.98 & \textbf{1.00} & \textbf{0.99} \\
 & replaced   & -  & - & - & & -  & - & - & & \textbf{1.00} & \textbf{0.40} & \textbf{0.57} & & 0.38 & 0.40 & 0.39 & & \textbf{1.00} & \textbf{0.40} & \textbf{0.57} \\
 & swapped   & -  & - & - & & 0.57  & \textbf{0.50} & \textbf{0.53} & & \textbf{1.00} & 0.04 & 0.07  & & 0.07 & 0.11 & 0.09 & & \textbf{1.00} & 0.11 & 0.19 \\ \cline{2-21}
\parbox[t]{4mm}{\multirow{-6}{*}{\rotatebox[origin=c]{90}{\textbf{P2P}}}}& conforming & 0.00  & 0.00 & 0.00 & & 0.87  & \textbf{1.00} & 0.93 & & 0.85 & \textbf{1.00} & 0.92 & & 0.97 & 0.83 & 0.89 & & \textbf{0.98} & \textbf{1.00} & \textbf{0.99} \\ \bottomrule

\end{tabular}}
\vspace{-1em}
\end{table}

\myparit{Overall Results} Our approach achieves a relatively high F1-score for all deviation pattern types and conforming traces across all used event logs. Thereby, it accurately detects repeated activities, likely because these are recognizable with only a general idea of what repetitions are. 
Insertions are also detected reasonably well, whereas skips and replacements are detected less consistently. 
We suspect that these are harder to detect within many possibly missing activities, whereas inserted activities must be observed in the trace, leading to fewer possibly inserted activities. 
Further, our approach struggles to detect swaps. The reason could be that capturing an incorrect order is more challenging as this can be specific to the process. Our approach detects conforming traces accurately, reaching an F1-score above 0.85. 
In most cases, it achieves higher precision than recall, meaning that not all deviations are detected but there are few false alarms.

\myparit{xSemAD \cite{busch2024xsemad}} xSemAD detects a large amount of deviations that constitute false positives, as indicated by the low precision. This also leads to low recall and precision of conforming traces. That is since it outputs declarative constraint violations rather than the five deviation patterns directly. Consequently, not traces but pairs of activities are analyzed. Thus, one wrong constraint can lead to many false positives. This shows that the goal of xSemAD differs significantly from the goal of our approach. Also, this indicates that traces should be analyzed individually for accurate feedback on trace-level.

\myparit{DABL \cite{guan2024dabl}} This fine-tuning strategy is often effective in detecting insertions and repetitions. However, skips are rarely recognized. We suspect that this is since the fine-tuned model overfits skipped activities in the training data and is not able to generalize towards unseen processes. Notably, the F1 of swaps is similar to our approach and higher for P2P and SAP-SAM, indicating that fine-tuning can help to detect re-ordered activities. 

\myparit{No Log Context} In contrast to our approach, this baseline achieves lower F1 scores across all deviation pattern types. Especially skipped and replaced activities are not detected accurately. This is because it is nearly impossible to detect skipped activities with no idea which activities could have occurred. The only pattern type that is recognized often is \textit{repeated}, likely since the LLM has some encoded idea of what repetitions are. 

\myparit{Log Context} Only supplying the general-purpose LLM with log context achieves relatively high F1-scores across all deviation pattern types. For skips in SAP-SAM and PROP, the F1 is even higher than our approach. Together with better accuracy than the other baselines, this indicates that log context already helps to detect undesired behavior. However, for P2P, the performance is significantly worse than our approach. We believe that only log context is not sufficient in more complex processes that contain meaningful rather than randomized deviations, such as P2P.

\begin{table}[h]
\vspace{-1.5em}
\centering
\caption{Standard deviation of precision, recall, and F1 for P2P over three random seeds.}
\label{tab:robust}
\resizebox{\textwidth}{!}{
\begin{tabular}{l>{\centering\arraybackslash}m{.75cm}>{\centering\arraybackslash}m{.75cm}>{\centering\arraybackslash}m{.75cm}m{0.05cm}>{\centering\arraybackslash}m{.75cm}>{\centering\arraybackslash}m{.75cm}>{\centering\arraybackslash}m{.75cm}m{0.05cm}>{\centering\arraybackslash}m{.75cm}>{\centering\arraybackslash}m{.75cm}>{\centering\arraybackslash}m{.75cm}m{0.05cm}>{\centering\arraybackslash}m{.75cm}>{\centering\arraybackslash}m{.75cm}>{\centering\arraybackslash}m{.75cm}m{0.05cm}>{\centering\arraybackslash}m{.75cm}>{\centering\arraybackslash}m{.75cm}>{\centering\arraybackslash}m{.75cm}}
\toprule
\multirow{3}{*}{\textbf{Pattern}} & \multicolumn{7}{c}{\textbf{Fine-Tuned Approaches}} & &
 \multicolumn{7}{c}{\textbf{Vanilla LLM}} & &
 \multicolumn{3}{c}{\multirow{2}{*}{\textbf{Our Approach}}} \\ 
 & \multicolumn{3}{c}{\textbf{xSemAD}\cite{busch2024xsemad}} & & \multicolumn{3}{c}{\textbf{DABL}\cite{guan2024dabl}} &  & \multicolumn{3}{c}{\textbf{No Log Context}}&  & \multicolumn{3}{c}{\textbf{Log Context}} &  & \multicolumn{3}{c}{}  \\
  & \textbf{Prec.}& \textbf{Rec.} & \textbf{F1} & & \textbf{Prec.} & \textbf{Rec.} & \textbf{F1} & & \textbf{Prec.} & \textbf{Rec.} & \textbf{F1} & & \textbf{Prec.} & \textbf{Rec.} & \textbf{F1} & & \textbf{Prec.} & \textbf{Rec.} & \textbf{F1} \\ \midrule

 inserted  & ± .04 & ± .01 & ± .05 & & ± .00   & ± .00   & ± .00  &  & ± .00   & ± .00   & ± .00  &  & ± .24   & ± .00   & ± .04  &  & ± .00   & ± .05   & ± .04   \\
 skipped    & ± .02 & ± .07 & ± .03& & ± .00   & ± .00   & ± .00  &  & ± .47 & ± .07 & ± .12  &  & ± .04   & ± .02   & ± .03 &  & ± .00   & ± .00 & ± .00 \\
 repeat  & -      & -      & -      & & ± .00   & ± .00   & ± .00  &  & ± .02   & ± .02   & ± .01 &   & ± .01 & ± .06 & ± .03 & & ± .00 & ± .00 & ± .00 \\
 replace & -      & -      & -      & & -      & -      & -     &  & ± .14   & ± .00 & ± .03   &  & ± .14   & ± .00   & ± .05 & & ± .00 & ± .00 & ± .00 \\
 swap    & -      & -      & -     &  & ± .00   & ± .00   & ± .00 &   & ± .00   & ± .00 & ± .00   &  & ± .01   & ± .08   & ± .03 &  & ± .00   & ± .00 & ± .00 \\ \cline{1-20}
conforming  & ± .00   & ± .00   & ± .00  &  & ± .00   & ± .00   & ± .00 &   & ± .00   & ± .00   & ± .00   &  & ± .00   & ± .03   & ± .01  &  & ± .00 & ± .00   & ± .00 \\ \bottomrule

\end{tabular}}
\end{table}

\mypar{Robustness} To evaluate how robust the approaches perform across multiple runs, we apply them three times with different random seeds to the P2P log. \autoref{tab:robust} shows the standard deviation of all metrics over these three runs. We see that DABL performs most consistently; our approach shows some variety in recall of insertions (± 0.05). Both Vanilla LLMs differ considerably across the three runs, especially in precision. Across all three runs, our approach outperforms all baselines. 
This indicates that RAG both improves performance and reduces variability in the results.

\mypar{Computational efficiency} We assess computational efficiency based on the time for training (if required) and inference. xSemAD and DABL require training in form of LLM fine-tuning, whereas our approach only requires the population of the knowledge base in the offline component. The Vanilla LLMs are used without any training. 
Since we did not train xSemAD and DABL ourselves, we report the training times from the original papers, which used considerably stronger GPUs than we had available.

\autoref{tab:eva_time} shows the training time and average inference time per event log. Training of xSemAD took 227 hours (more than 9 days), which is even surpassed by DABL with 267 hours. In contrast, our approach only required 1.30 hours for the knowledge base population. This difference is significant, especially considering that we used less computational resources. xSemAD and DABL used more process models during training than our approach, leading to an increase in training time that is not justified by a better performance. 
For the average inference time, xSemAD is the quickest by far. This is because not every single trace of the event log is checked by the LLM. Rather, the LLM uses all activities to generate a set of constraints once, after which constraint fulfillment is checked per trace. Although this decreases inference time, the accuracy is impacted heavily. DABL requires the most inference time due to its LLM with a size of around 30 GB. Our approach takes less time than DABL but more than the Vanilla LLM baselines. 
This inference time of our approach
can be considered long, but our experiments showed that each trace has to be analyzed individually for accurate performance.

\begin{table}[t]
\centering
\caption{Training and average inference time for used datasets.}
\label{tab:eva_time}
\resizebox{\textwidth}{!}{
\begin{tabular}{m{2.7cm}|rm{0.15cm}rm{0.15cm}rm{0.15cm}rm{0.15cm}r}
\toprule
 &\textbf{xSemAD}\cite{busch2024xsemad} & & \textbf{DABL \cite{guan2024dabl}} & & \textbf{No Log Context} & & \textbf{Log Context} & & \textbf{Our Approach} \\ \midrule
\textbf{Training (h)}    & 227.00 & & 267.51 & & - & & -  &  & 1.30 \\
\textbf{Avg. inference (min)} & 0.40 &  & 22.50 && 4.01 & & 4.11 & & 9.30 \\ \bottomrule 
\end{tabular}}
\vspace{-1em}
\end{table}

\subsection{Real-Life Application} \label{sec:eva_real}
In this section, we demonstrate the practical applicability of our approach by showing which undesired behavior it detects in a well-known real-life event log. 
To this end, we apply it to the event log from the BPI Challenge 2019, focusing on the consignment sub-process\footnote{\url{https://doi.org/10.4121/uuid:d06aff4b-79f0-45e6-8ec8-e19730c248f1}}. This purchase order handling process only contains activities related to the ordering and receiving of goods, whereas payment is handled separately. The three most frequent traces that the approach provides to the LLM as part of our prompt, are:
\begin{compactenum}[(1)]
\item $\langle$Create Purchase Order (PO) Item, Record Goods Receipt$\rangle$
\item $\langle$Create Purchase Requisition (PReq) Item, Create PO Item, Record Goods Receipt$\rangle$
\item $\langle$Create PO Item, Record Goods Receipt, Record Goods Receipt$\rangle$
\end{compactenum}

Since true deviations are unknown, we conducted a qualitative assessment, similar to other works \cite{van2021natural,caspary2023does,busch2024xsemad}. 
In particular, we illustrate our approach based on three traces $t_1$, $t_2$, and $t_3$, which we show in \autoref{tab:bpic19}.
For each of these traces, the table displays the trace itself, one exemplary retrieved trace from the knowledge base including the corresponding retrieved deviation, and the output of our approach.

\begin{table}[ht]
\centering
\caption{Exemplary illustration of our approach in BPI Challenge 2019. Shown are three analyzed traces, one retrieved trace-deviation pair, and the output of our approach.} \label{tab:bpic19}
\begin{threeparttable}
\renewcommand{\TPTminimum}{\linewidth}
\makebox[\linewidth]{%
\resizebox{\textwidth}{!}{
\begin{tabular}{m{.3cm}m{2.8cm}m{.1cm}m{3.8cm}m{.1cm}m{2.9cm}m{.1cm}m{2.9cm}}
\toprule
\multirow{2}{*}{\textbf{Trace}} &&&  \multicolumn{3}{c}{\textbf{Retrieved Example from Knowledge Base}} & &\multirow{2}{*}{\textbf{Approach Output}}\\ \cline{4-6}
&&& \textbf{Retrieved Trace} && \textbf{Retrieved Deviation} && \\\midrule
$t_1$&$\langle$Create PO Item, Change Quantity, Delete PO Item$\rangle$ && $\langle$Receive Order, Check Order, \dots , Receive Goods$\rangle$ && No Deviation. && Replace(['Record GR'], ['Delete PO Item'])  \\ \midrule
$t_2$&$\langle$Create PO Item, Change Storage Location, Record GR$\rangle$ && $\langle$Create PO Item', 'Record Invoice Receipt$\rangle$ && Skipped: 'Receive Goods' && Insert(['Change Storage Location'])  \\  \midrule
$t_3$&$\langle$Create PO Item, Change Quantity, Change Quantity, Record GR$\rangle$ && $\langle$Create PO, Print and Send PO, Change Price, Change Price, Receive Goods, Scan Invoice, Book Invoice$\rangle$ && Repeated: 'Change Price' && Repeat(['Change Quantity'])    \\\bottomrule
\end{tabular}
}}

\vspace{-.5em}
\begin{tablenotes}
\sffamily
\medskip
\item PO = Purchase Order; GR = Goods Receipt
\end{tablenotes}
\end{threeparttable}
\vspace{-1em}
\end{table}

In the first trace $t_1$, our approach detects a replacement of \emph{Record GR}---which occurs in the three most frequent traces---by \textit{Delete PO Item} at the end of the trace. This matches the assumption that deleting an order item has negative impact on the process goal to successfully complete the order and is consequently undesired. The retrieved trace ends with \emph{Receive Goods}, which is similar to the replaced activity. 
For trace $t_2$, our approach recognizes a problem with \emph{Change Storage Location}, which is inserted according to the output. This aligns with the expectation that changing the storage location is probably inefficient and hence undesired. The insertion might have been detected with the help of the three most frequent traces, in which \emph{Change Storage Location} does not occur.
In the last trace $t_3$, our approach detects a repetition of \emph{Change Quantity} as it occurs twice, indicating undesired inefficiency. The retrieved trace has a similar deviation as \emph{Change Price} is repeated.  For all traces, the retrieved examples contain similar behavior. In summary, this analysis reveals insights into the process flow of purchasing goods, highlighting undesired behavior with potential impact on the organization.

\section{Conclusion}\label{sec:conclusion}
In this paper, we proposed an approach for detecting undesired behavior in event logs by means of RAG. 
The approach does not rely on any fine-tuning but rather utilizes the capabilities of a general-purpose LLM, supplied by means of RAG with relevant examples from other processes as well as dynamically selected context from the event log. 
Our experiments demonstrate that this approach not only significantly reduces computational resource requirements but also enhances performance compared to plain fine-tuning. In particular, the additional log context our approach provides to the LLM was essential for detecting missing activities. In addition, RAG improved the results, especially in more complex processes, indicating that it is possible to generalize behavior from other processes. 
Finally, the output of our approach is able to capture deviations that involve any number of activities rather than constraint violations which refer to at most two activities such as xSemAD. 

Our approach is subject to limitations. 
First, our approach returns slightly different answers across multiple random seeds, leading to variety in its performance. Although this is undesired, our approach has been shown to outperform the baselines, meaning that the variety is not negatively impacting the performance to an extend where our approach performs worse. 
Second, our approach extracts explicit information from the event log, assuming that frequent activities and frequent traces help to identify undesired behavior. Although this assumption turned out to be true in the used experiments, it might not hold for all cases.  
We acknowledge that the findings for our generated logs as well as the manually designed process from \cite{hosseinpour2024auditors} might not be generally applicable. 
Third, swaps are not accurately detected by our approach, potentially because re-ordered activities are hard to detect for LLMs. Moreover, our experiments showed that swaps are the only pattern type where DABL performs as good or better than our approach, indicating that fine-tuning on particular process behaviors could be helpful. 
Fourth, our approach does require process models to be available for the offline component. Thus, users of the approach must have access to such process models, which can be done based on, e.g., a suitable subset of SAP-SAM. To achieve best performance, these underlying processes should be similar to the investigated ones. However, this is not a pre-requisite as our approach has been shown to work without ensuring process similarity. Future works might investigate the effects of different model collections in the offline component.
Fifth, real-world processes might be customized in detail. Thus, knowledge from other processes could not be applicable out of the box. In these cases, organization-specific adaption could help to incorporate the process customization.
Last, the inference time of our approach per event log is relatively long, possibly decreasing practical applicability. However, only this strategy has proven to achieve accurate performance, especially in contrast xSemAD which has a lower inference time but also many false positive detections. 

Our approach represents a novel direction in the application of LLMs in process mining research, being the first to employ RAG as a means to provide LLMs access to process knowledge to solve a process mining task. 
In the future, we want to investigate whether a more sophisticated retrieval of information from other processes might improve the performance. For example, we could experiment both with more sophisticated trace embeddings as well as embeddings and retrieval of not just traces but additional information from the process model collection.
This might help to extend our approach towards undesired behavior from perspectives other than the control-flow, to which our current approach is limited to.
Also, the inference time of our approach might be reduced with the batching of traces, utilizing the full allowed input tokens of the LLM.
Further, given the variety of process models in SAP-SAM, a more detailed investigation of these models such as a cross-fold validation could help to show the robustness of our approach.
In addition, we want to investigate how imbalanced processes with significantly more desired than undesired traces influence our approach, a setting more likely in reality \cite{grohs2025proactive}. In such cases, it could be even more helpful to include log context into the prompt as the most frequent traces represent good references for desired behavior. For the current experiments, we created for a balanced dataset to investigate many different and diverse undesired executions.
Further, we want to assess whether a combination of fine-tuning, RAG, and the inclusion of explicit context from the event log might result in more accurate detections. Our findings suggest that this might be a valuable direction which could benefit from the advantages of these three ideas in conjunction. Using the insights gained through our work, we aim to also employ LLMs in combination with RAG in other process mining tasks. For instance, process improvement opportunities could be derived based on access to knowledge of similar processes. 

%
%
%
\bibliographystyle{splncs04}
\bibliography{sample}

\begin{thebibliography}{10}
\providecommand{\url}[1]{\texttt{#1}}
\providecommand{\urlprefix}{URL }
\providecommand{\doi}[1]{https://doi.org/#1}

\bibitem{van2021natural}
van~der Aa, H., Rebmann, A., Leopold, H.: Natural language-based detection of semantic execution anomalies in event logs. Inf Syst  \textbf{102},  101824 (2021)

\bibitem{van2012makes}
van~der Aalst, W.M.: What makes a good process model? lessons learned from process mining. SoSyM  \textbf{11}(4),  557--569 (2012)

\bibitem{bernardi2024conversing}
Bernardi, M.L., Casciani, A., Cimitile, M., Marrella, A.: Conversing with business process-aware large language models: the {BPLLM} framework. J Intell Inf Syst pp. 1--23 (2024)

\bibitem{böhmer2017multi}
B{\"o}hmer, K., Rinderle-Ma, S.: Multi instance anomaly detection in business process executions. In: BPM. pp. 77--93. Springer (2017)

\bibitem{böhmer2020association}
B{\"o}hmer, K., Rinderle-Ma, S.: Mining association rules for anomaly detection in dynamic process runtime behavior and explaining the root cause to users. Inf Syst  \textbf{90},  101438 (2020)

\bibitem{busch2024xsemad}
Busch, K., Kampik, T., Leopold, H.: xsemad: Explainable semantic anomaly detection in event logs using sequence-to-sequence models. In: BPM. pp. 309--327. Springer (2024)

\bibitem{carmona2018conformance}
Carmona, J., van Dongen, B., Solti, A., Weidlich, M.: Conformance checking. Springer (2018)

\bibitem{caspary2023does}
Caspary, J., Rebmann, A., van~der Aa, H.: Does this make sense? machine learning-based detection of semantic anomalies in business processes. In: BPM. pp. 163--179. Springer (2023)

\bibitem{Dunzer_2019SOA_CC}
Dunzer, S., Stierle, M., Matzner, M., Baier, S.: Conformance checking: A state-of-the-art literature review. In: S-BPM ONE. p. 1–10. {ACM} (2019)

\bibitem{estrada2024mapping}
Estrada-Torres, B., del R{\'i}o-Ortega, A., Resinas, M.: Mapping the landscape: Exploring large language model applications in business process management. In: BPMDS. pp. 22--31. Springer (2024)

\bibitem{friedrich2011process}
Friedrich, F., Mendling, J., Puhlmann, F.: Process model generation from natural language text. In: CAISE. pp. 482--496. Springer (2011)

\bibitem{garcia_2017_complete}
Garc{\'\i}a-Ba{\~n}uelos, L., Van~Beest, N., Dumas, M., La~Rosa, M., Mertens, W.: Complete and interpretable conformance checking of business processes. Trans Softw Eng  \textbf{44}(3),  262--290 (2017)

\bibitem{grohs2024beyond}
Grohs, M., van~der Aa, H., Rehse, J.R.: Beyond log and model moves in conformance checking: Discovering process-level deviation patterns. In: BPM. pp. 381--399. Springer (2024)

\bibitem{grohs2025proactive}
Grohs, M., Pfeiffer, P., Rehse, J.R.: Proactive conformance checking: An approach for predicting deviations in business processes. Inf Sys  \textbf{127},  102461 (2025), \url{https://www.sciencedirect.com/science/article/pii/S0306437924001194}

\bibitem{guan2024dabl}
Guan, W., Cao, J., Gao, J., Zhao, H., Qian, S.: Dabl: Detecting semantic anomalies in business processes using large language models (2024), \url{https://arxiv.org/abs/2406.15781}

\bibitem{hosseinpour2024auditors}
Hosseinpour, M., Jans, M.: Auditors’ categorization of process deviations. J Inf Sys  \textbf{38}(1),  67--89 (2024)

\bibitem{lahann2022lstm}
Lahann, J., Pfeiffer, P., Fettke, P.: Lstm-based anomaly detection of process instances: benchmark and tweaks. In: ICPM Workshops. pp. 229--241. Springer (2022)

\bibitem{deleoni2012aligning}
de~Leoni, M., Maggi, F.M., van~der Aalst, W.M.P.: Aligning event logs and declarative process models for conformance checking. In: BPM. pp. 82--97. Springer (2012)

\bibitem{lewis2020retrieval}
Lewis, P., Perez, E., Piktus, A., Petroni, F., Karpukhin, V., Goyal, N., K{\"u}ttler, H., Lewis, M., Yih, W.t., Rockt{\"a}schel, T., et~al.: Retrieval-augmented generation for knowledge-intensive nlp tasks. Adv Neural Inf Process Sys  \textbf{33},  9459--9474 (2020)

\bibitem{monti2024nl2processops}
Monti, F., Leotta, F., Mangler, J., Mecella, M., Rinderle-Ma, S.: Nl2processops: towards llm-guided code generation for process execution. In: International Conference on Business Process Management. pp. 127--143. Springer (2024)

\bibitem{Mun16fundamentalsCC}
Munoz-Gama, J.: Conformance Checking and Diagnosis in Process Mining: Comparing Observed and Modeled Processes. Springer (2016)

\bibitem{nolle_binet}
Nolle, T., Luettgen, S., Seeliger, A., Mühlhäuser, M.: Binet: Multi-perspective business process anomaly classification. Inf Syst  \textbf{103},  101458 (2019)

\bibitem{nussbaum2024nomic}
Nussbaum, Z., Morris, J.X., Duderstadt, B., Mulyar, A.: Nomic embed: Training a reproducible long context text embedder  (2024), \url{https://arxiv.org/abs/2402.01613}

\bibitem{rebmann2024evaluating}
Rebmann, A., Schmidt, F.D., Glavaš, G., van Der~Aa, H.: Evaluating the ability of llms to solve semantics-aware process mining tasks. In: ICPM. pp. 9--16 (2024)

\bibitem{sola2022sap}
Sola, D., Warmuth, C., Sch{\"a}fer, B., Badakhshan, P., Rehse, J.R., Kampik, T.: {SAP Signavio Academic Models}: a large process model dataset. In: ICPM Workshops. pp. 453--465. Springer (2022)

\bibitem{team2024gemma}
Team, G.: Gemma2  (2024). \doi{10.34740/KAGGLE/M/3301}, \url{https://www.kaggle.com/m/3301}

\end{thebibliography}

\end{document}